\documentclass{bmvc2k}

\title{Deep Structured Models For Group Activity Recognition}

\addauthor{Zhiwei Deng}{zhiweid@sfu.ca}{1}
\addauthor{Mengyao Zhai}{mzhai@sfu.ca}{1}
\addauthor{Lei Chen}{chenleic@sfu.ca}{1}
\addauthor{Yuhao Liu}{yla305@sfu.ca}{1}
\addauthor{Srikanth Muralidharan}{smuralid@sfu.ca}{1}
\addauthor{Mehrsan Javan Roshtkhari}{mehrsan@sportlogiq.com}{2}
\addauthor{Greg Mori}{mori@cs.sfu.ca}{1}

\addinstitution{
 School of Computing Science\\
 Simon Fraser University\\
 Burnaby, BC, Canada
}
\addinstitution{
 SPORTLOGiQ\\
 Montreal, QC, Canada
}

\runninghead{Student, Prof, Collaborator}{Deep Structured Models}

\def\etal{\emph{et al}\bmvaOneDot}

\begin{document}

\maketitle

\begin{abstract}
  This paper presents a deep neural-network-based hierarchical graphical model for individual and group activity recognition in surveillance scenes.  Deep networks are used to recognize the actions of individual people in a scene.  Next, a neural-network-based hierarchical graphical model refines the predicted labels for each class by considering dependencies between the classes.  This refinement step mimics a message-passing step similar to inference in a probabilistic graphical model.  We show that this approach can be effective in group activity recognition, with the deep graphical model improving recognition rates over baseline methods.
\end{abstract}

%------------------------------------------------------------------------- 
\section{Introduction}
Event understanding in videos is a key element of computer vision systems in the context of visual surveillance, human-computer interaction, sports interpretation, and video search and retrieval.  Therefore events, activities, and interactions must be represented in such a way that retains all of the important visual information in a compact and rich structure. Accurate detection and recognition of atomic actions of each individual person in a video is the primary component of such a system, and also the most important, as it affects the performance of the whole system significantly. Although there are many methods to determine human actions in uncontrolled environments, this task remains a challenging computer vision problem, and robust solutions would open up many useful applications.  The standard and yet state-of-the-art pipeline for activity recognition and interaction description consists of extracting hand-crafted local feature descriptors either densely or at a sparse set of interest points (e.g., HOG, MBH, ...) in the context of a Bag of Words model~\cite{WangS_ICCV13}. These are then used as the input either to a discriminative or a generative model. In recent years, it has been shown that deep learning techniques can achieve state-of-the-art results for a variety of computer vision tasks including action recognition~\cite{SimonyanZ_NIPS14, Karpathy_CVPR2014}.

On the other hand, understanding of complex visual events in a scene requires exploitation of richer information rather than individual atomic activities, such as recognizing local pairwise and global relationships in a social context and interaction between individuals and/or objects~\cite{lan12_cvpr, ramanathan13_cvpr, ZhuNR13, RyooA11, BrendelT11}. This complex scene description remains an open and challenging task. It shares all of difficulties of action recognition, interaction modeling\footnote{The term ``interaction'' refers to any kind of interaction between humans, and humans and objects that are present in the scene, rather than activities which are performed by a single subject.}, and social event description. Formulating this problem within the probabilistic graphical models framework provides a natural and powerful means to incorporate the hierarchical structure of group activities and interactions~\cite{Tian_NIPS2010, lan12_cvpr}. Given the fact that deep neural networks can achieve very competitive results on the single person activity recognition tasks, they can, produce better results when they are combined with other methods, e.g. graphical models, in order to capture the dependencies between the variables of interest~\cite{TompsonJLB14}. Following a similar idea of incorporating spatial dependency between variables into the deep neural network in a joint-training process presented~\cite{TompsonJLB14}, here we focus on learning interactions and group activities in a surveillance scene by employing a graphical model in a deep neural network paradigm. 

\begin{figure*}
	\centerline{\includegraphics[width=0.7\textwidth]{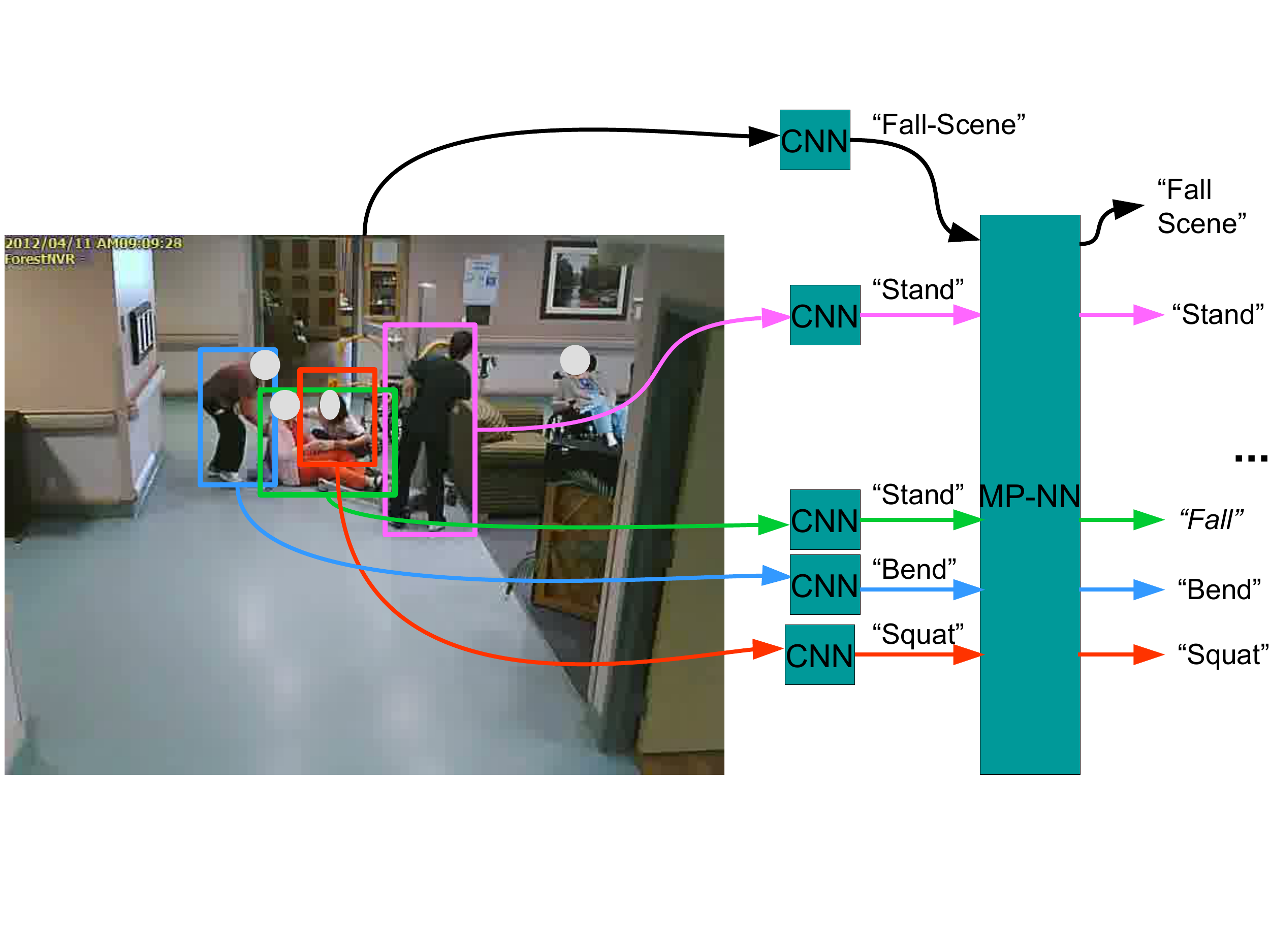}}
	\caption{Recognizing individual and group activities in a deep network.  Individual action labels are predicted via CNNs.  Next, these are refined through a message passing neural network which considers the dependencies between the predicted labels.}
	\label{fig:algorithm_overview}
\end{figure*}

In this paper, our main goal is to address the problem of \emph{group activity understanding} and \emph{scene classification} in complex surveillance videos using a deep learning framework. More specifically, we are focused on learning individual activities and describing the scene simultaneously while considering the pair-wise interactions between individuals and their global relationship in the scene. This is achieved by combining a Convolutional Network (ConvNet) with a probabilistic graphical model as additional layers in a deep neural network architecture into a unified learning framework. The probabilistic graphical models can be seen as a refining process for predicting class labels by considering dependencies between individual actions, body poses, and group activities. The probabilistic graphical model is modeled by a multi-step message passing neural network and the predicted label refinement is carried out through belief propagation layers in the neural network. \figurename~\ref{fig:algorithm_overview} depicts an overview of our approach for label refinement. Experimental results show the effectiveness of our algorithm in both activity recognition and scene classification. 

%The paper is organized as follows: 

%
%The contributions of this paper can be summarized as follows: (i) we propose a combined CNN and graphical model by approximating a variant of the message passing algorithm... 
%[[contribution: 1. combine graphical model and CNN by mimicking the message passing algorithm. 2. traditional CNN learns low level features, we propose a method to learn high-level semantic features that can be used for scene classification. 3. Applying deep learning in surveillance video]]
%We show that the combination and joint training of these two models improves performance, and allows us to significantly outperform existing state-of-the-art models on the task of group activity recognition
%Therefore, and our objectives are to (i) to describe all the individual activities in a scene, (ii) recognize and classify pair-wise interaction between individuals, and individuals/groups of individuals.  

%**************** Figures to be added *****************
%[[ We may add a figure: Figure ~\ref{} shows an example of a complex visual event in a nursing care institution. There are lots of interactions between individuals that have to be described in order to create a complex event description algorithm. ]]

\vspace*{-1.5mm}

\begin{figure*}
	\centerline{\includegraphics[width=0.7\textwidth]{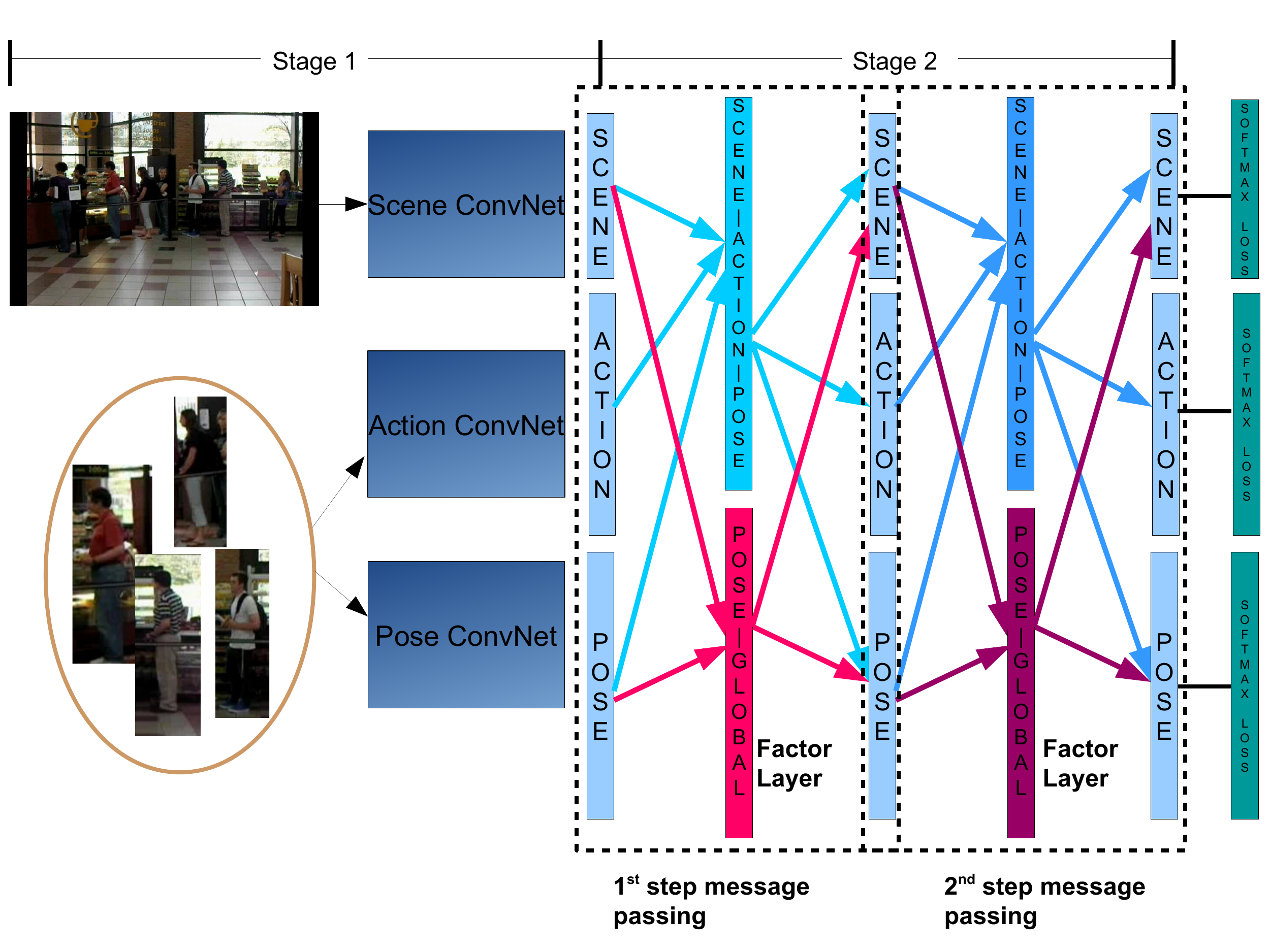}}
	\caption{A schematic overview of our message passing CNN framework. Given an image frame and the detected bounding boxes around each person, our model predicts scores for individual actions and the group activities. The predicted labels are refined by applying a belief propagation-like neural network. This network considers the dependencies between individual actions and body poses, and the group activity. The model learns the message passing parameters and performs inference and learning in unified framework using back-propagation.}
	\label{fig:label_refinement_procedure}
\end{figure*}

%------------------------------------------------------------------------------------------
\section{Previous Work}
\label{sec:pw}

The analysis of human activities is an active area of research. Decades of research on this topic have produced a diverse set of approaches and a rich collection of activity recognition algorithms. Readers can refer to recent surveys such as Poppe~\cite{Poppe10} and Weinland et al.~\cite{WeinlandRB10} for a review. 
Many approaches concentrate on an activity performed by a single person, including state of the art deep learning approaches~\cite{SimonyanZ_NIPS14, Karpathy_CVPR2014}. 

In the context of scene classification and group activity understanding, many approaches use a hierarchical representation of activities and interactions for collective activity recognition~\cite{lan12_cvpr}. They have been focused on capturing spatio-temporal relationships between visual cues either by imposing a richer feature descriptor which accounts for context~\cite{TranGKS2014, ChoiSS2009} or a context-aware inference mechanism~\cite{ChoiS12, AmerLT_ECCV2014}. Hierarchical graphical models~\cite{RyooA11, LanWYRM12, lan12_cvpr, AmerLT_ECCV2014}, AND-OR graphs~\cite{AmerXZTZ12, gupta09_cvpr}, and dynamic Bayesian networks~\cite{ZhuNR13} are among the representative approaches for group activity recognition.

In traditional approaches, local-hand crafted features/descriptors has been employed to recognize atomic activities. Recently, it has been shown that the use of deep neural networks can by itself outperform other algorithms for atomic activity recognition. However, no prior art in the CNN-based video description used activities and scene information jointly in a unified graphical representation for scene classification. Therefore, the main objective of this research is to develop a system for activity recognition and scene classification which simultaneously uses the action and scene labels in a neural network-based graphical model to refine the predicted labels via a multiple-step message passing.

More closely related to our approach is work combining graphical models with convolutional neural networks~\cite{TompsonJLB14, deng2014ECCV}. In~\cite{TompsonJLB14}, a one step message passing is implemented as a convolution operation in order to incorporate spatial relationship between local detection responses for human body pose estimation. In another study, Deng \etal~\cite{deng2014ECCV} propose an interesting solution to improve label prediction in large scale classification by considering relations between the predicted class labels. They employ a probabilistic graphical model with hard constraints on the labels on top of a neural network in a joint training process. In essence, our proposed algorithm follows a similar idea of considering dependencies between predicted labels for the actions, group activities, and the scene label to solve the group activity recognition problem. Here we focus on incorporating those dependencies by implementing the label refinement process via an inter-activity neural network, as shown in \figurename{~\ref{fig:label_refinement_procedure}}. The network learns the message passing procedure and performs inference and learning in unified framework using back-propagation.

%Actions, representations:
%
%Morariu et al.~\cite{morariu11eventstructure}
%Amer et al.~\cite{AmerXZTZ12}
%Gupta et al.~\cite{gupta09_cvpr}
%Lan et al.~\cite{lan12_cvpr}, Ramanathan et al.~\cite{ramanathan13_cvpr}
%Zhu et al.~\cite{ZhuNR13}
%Laptev work on movie scripts (CVPR08 old, ECCV14 recent)  
%Marszalek et al.~\cite{marszalek09_cvpr}
%Bojanowski et al.~\cite{BojanowskiLBLPSS14}
%
%Intille and Bobick~\cite{IntilleB01}, Medioni et
%al.~\cite{MedioniCBHN01}, Moore et al.~\cite{moore99_iccv}
%
%Khamis et al.~\cite{khamis-eccv2012}
%

%Knowledge representation, structured queries
%Lan et al.~\cite{lan12_eccv}, Siddiquie et al.~\cite{siddiquie11_cvpr}
%Zhu et al.~\cite{ZhuFF14}
%Zitnick et al.~\cite{ZitnickPV13}, common sense knowledge
%Descriptions, e.g. Baby talk work~\cite{KulkarniPODLCBB13}
%General need for more complex representation power, a trend of
%approaches revisiting and enhancing this line of AI work.

%---------------------------------------------------------------
%new references:
%tompson's NIPS 2014. ~\cite{TompsonJLB14}  --- most similar one, but only simply using spatial information to filter out false positives
%~\cite{rosslearning} --- exploring other variants of message passing
%~\cite{Tian_NIPS2010} --- group activity understanding
%~\cite{ChoiS12} --- group activity understanding
%Mohhammad Amer \cite{AmerLT_ECCV2014}, eccv 2012
%Jia Deng eccv 2014 --- simple way to combine graphical model and CNN by stacking graphical model and CNN
%~\cite{AlexNIPS2012} --- AlexNet

%------------------------------------------------------------------------------------------------------------------
\section{Model}
Considering the architecture of our proposed structured label refinement algorithm for group activity understanding (see \figurename~\ref{fig:label_refinement_procedure}), the key part of the algorithm is a multi-step message passing neural network. In this section, we describe how to combine neural networks and graphical models by mimicking a message passing algorithm and how to carry out the training procedure.

%In this section, we explain our model in the following sections: (1) combining neural network and graphical model by mimicking message passing algorithm, (2) different components in message passing CNN for recognizing group activities and individual actions, (3) training graphical model in neural network by learning multi-step message passing, and corresponding implementation details.

\subsection{Graphical Models in a Neural Network}
\label{sec:mp}

Graphical models provide a natural way to hierarchically model group activities and capture the semantic dependencies between group and individual activities~\cite{Tian_NIPS2010}. A graphical model defines a joint distribution over states of a set of nodes. For instance, one can use a factor graph, in which each $\phi_{i}$ corresponds to a factor over a set of related variable nodes $x_{i}$ and $y_{i}$, and models interactions between these nodes in a log-linear fashion:
\begin{equation}
P(X,Y) \propto \prod_{i}{\phi_{i}(x_{i},y_{i})} \propto exp{(\sum_{k}{w_{k}f_k(x,y)})}   
\label{eq:graphical_model}         
\end{equation}
where $X$ are the inputs and $Y$ the predicted labels, with weighted ($w_{k}$) feature functions $f_k$. 

When performing inference in a graphical model, belief propagation is often adopted as a way to infer states or probabilities of variables. In the belief propagation algorithm, each step of message passing first collects relevant information from connected nodes to a factor node, which represents the joint distribution (dependencies) over states, then passes these messages to variable nodes by marginalizing over states of irrelevant variables.  

\begin{figure*}
	\centerline{\includegraphics[width=0.7\textwidth]{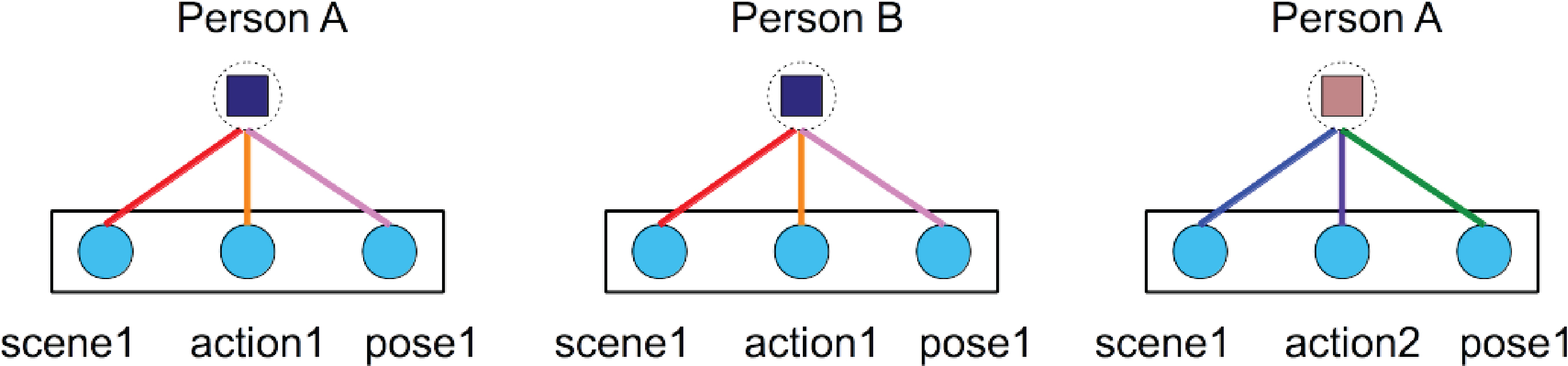}}
	\caption{Weight sharing scheme in neural network. We use a sparsely connected layer to represent message passing between variable nodes and factor nodes. Each factor node only connects to its relevant nodes. And factor nodes of same type share a template of parameters. For example, factor node 1 and 2 gathers information from a scene's scene1, a person's action1 and pose1, and share one template of parameters. And factor node 3 adopts another set of weights.}
\label{fig:weights}
\end{figure*}

Following this idea, we mimic the message passing process by representing each combination of states as a neuron in neural network, denoted as a ``factor neuron.''  While normal message passing calculates dependencies rigidly, a factor neuron can be used to learn and predict dependencies between states and pass messages to variable nodes.  In the setting of neural networks, this dependency representation becomes more flexible and can adopt varied types of neurons (linear, ReLU, Sigmoid, etc.). Moreover, by integrating graphical models into a neural network, the formulation of a graphical model allows for parameter sharing in the neural network, which not only reduces the number of free parameters to learn but also accounts for semantic similarities between factor neurons.  Fig.~\ref{fig:weights} shows the parameter sharing scheme for different factor neurons. 

%[[changed by zhiwei]]

\subsection{Message Passing CNN Architecture for Group Activity}

Representing group activities and individual activities as a hierarchical graphical model has proven successful~\cite{Tian_NIPS2010, ChoiS12, AmerXZTZ12}. We adopt a similar structured model that considers group activity, individual activity, and group-individual interactions together.
We introduce a new message passing Convolutional Neural Network framework as shown in Fig.~\ref{fig:label_refinement_procedure}. Our model has two main stages: (1) fine-tuned Convolutional Neural Networks that produce scene scores for a frame, and action and pose scores for each person in that frame; (2) Message Passing Neural Network phase capturing dependencies.

% from mengyao
Given an image $I$ and a set of person detections $\{I_{1}, I_{2},..., I_{M}\}$, the first stage of our model outputs raw scores of scene, action and poses for image $I$ and all detections ${I_{i}}$ in the image using fine-tuned CNNs. After a softmax normalization for each scene and person, these raw scores are taken as input of the graphical model part in the second stage. In the graphical model, outputs from CNNs correspond to unary potentials. Denote the scene-level, and per-person action and pose-level unary potentials for frame $I$ as $\mathbf{s}^{(0)}(I)$, $\mathbf{a}^{0}(I_{m})$, $\mathbf{r}^{(0)}(I_{m})$ respectively.  The superscript (0) is the index of message passing steps. We use $G$ to denote all group activity labels, $H$ to represent all the action labels and $Z$ to denote all the pose labels. Then the group activity in one scene can be represented as $g_{I}$, $\{h_{I_{1}},h_{I_{2}},...,h_{I_{M}}\}$, $\{z_{I_{1}},z_{I_{2}},...,z_{I_{M}}\}$ where $g_{I} \in G$ is the group activity label for image $I$, $h_{I_{i}}$ and $z_{I_{i}}$ are action labels and pose labels for a person $I_{m}$. 

Note that for training, the scene, action, and pose CNN models in stage 1 are fine-tuned from an AlexNet architecture pretrained using ImageNet data. The architecture is similar to that proposed by~\cite{AlexNIPS2012} for object classification with some minor differences such as pooling is done before normalization. The network consists of five convolutional layers followed by two fully connected layers, and a softmax layer that outputs individual class scores. We use the softmax loss, stochastic gradient descent and dropout regularization to train these three ConvNets.
%The inputs to the CNN are image patches of size $256*256$, and the outputs are probabilities over $N$ categories plus one background category. We use the softmax loss, stochastic gradient descent and dropout regularization to train these three ConvNets.
%%%%%%%%%%%%%%%%%%

In the second stage, we use the method mentioned in Sec.~\ref{sec:mp} to mimic message passing in a hierarchical graphical model for group activity in a scene. This stage can contain several steps of message passing. In each step, there are two types of passes: from outputs of step $k-1$ to factor layer and from factor layer to $k$ step outputs. In the $k^{th}$ message passing step, the first pass computes dependencies between states. The inputs to the $k^{th}$ step message passing are $\{s_{1}^{(k-1)}(I),...,s_{|G|}^{(k-1)}(I),a_{1}^{(k-1)}(I_{1}),...,a_{|H|}^{(k-1)}(I_{M}),r_{1}^{(k-1)}(I_{1}),...,r_{|Z|}^{(k-1)}(I_{M}))\}$, where 
$s_{g}^{(k-1)}(I)$ is the scene score of image $I$ for label $g$, $a_{h}^{(k-1)}(I_{m})$ is the action score of person $I_{m}$ for label $h$ and $r_{z}^{(k-1)}(I_{m}))$ is the pose score of person $I_{m}$ for label $z$. In the factor layer, the action, pose and scene interaction are calculated as: 
\begin{equation}
\phi_{j}(s_{g}^{(k-1)}(I),a_{h}^{(k-1)}(I_{m}),r_{z}^{(k-1)}(I_{m}))) = \mathbf{\alpha}_{g,h,z}[s_{g}^{(k-1)}(I),a_{h}^{(k-1)}(I_{m}),r_{z}^{(k-1)}(I_{m}))]^{T}
\end{equation}
where $\mathbf{\alpha}_{g,h,z}$ is a 3-d parameter template for combination of scene $g$, action $h$ and pose $z$.
Similarly, pose interactions for all people in the scene are calculated as:
\begin{equation}
\psi_{j}(s_{g}^{(k-1)}(I), \mathbf{r}) = \mathbf{\beta}_{tg}[s_{g}^{(k-1)}(I),  \mathbf{r}]^{T}
\end{equation}
where $\mathbf{r}$ is all output nodes for all people, $t$ is the factor neuron index for scene $g$. $T$ latent factor neurons are used for a scene $g$. Note that parameters $\alpha$ and $\beta$ are shared within factors that have the same semantic meaning. For the output of $k^{th}$ step message passing, the score for the scene label to be $g$ can be defined as:
\begin{equation}
s_{g}^{(k)}(I) = s_{g}^{(k-1)}(I) + \sum_{j \in \varepsilon_{1}^{s}}w_{ij}\phi_{j}(s_{g}^{(k-1)}(I),\mathbf{a},\mathbf{r};\mathbf{\alpha})) + \sum_{j \in \varepsilon_{2}^{s}}w_{ij}\psi_{j}(s_{g}^{(k-1)}(I), \mathbf{r};\mathbf{\beta})
\end{equation}
where $\varepsilon_{1}^{s}$ and $\varepsilon_{2}^{s}$ are the set of factor nodes that connected with scene $g$ in first factor component(scene-action-pose factor) and second factor component (pose-global factor) respectively.
Similarly, we also define action and pose scores after the $k^{th}$ message passing step as:
\begin{equation}
a_{h}^{(k)}(I_{m}) = a_{h}^{(k-1)}(I_{m}) + \sum_{j \in \varepsilon_{1}^{a}}w_{ij}\phi_{j}(a_{h}^{(k-1)}(I_{m}),\mathbf{s},\mathbf{r};\alpha)
\end{equation}

\vspace*{-3mm}

\begin{equation}
r_{z}^{(k)}(I_{m}) = r_{z}^{(k-1)}(I_{m}) + \sum_{j \in \varepsilon_{1}^{r}}w_{ij}\phi_{j}(r_{z}^{(k-1)}(I_{m}),\mathbf{a},\mathbf{s};\alpha) + \sum_{j \in \varepsilon_{2}^{r}}w_{ij}\psi_{j}(r_{z}^{(k-1)}(I_{m}), \mathbf{r};\mathbf{\beta})
\end{equation}
Note that $\varepsilon = \{\varepsilon_{1}^{s},\varepsilon_{2}^{s},\varepsilon_{1}^{a},\varepsilon_{1}^{r},\varepsilon_{2}^{r}\}$ are connection configurations in the pass from factor neurons to output neurons. These connections are simply the reverse of the configurations in the first pass, from input to factors. The model parameters $\{\mathbf{W},\mathbf{\alpha},\mathbf{\beta}\}$ are weights on the edges of the neural network. Parameter $\mathbf{W}$ represents the concatenation of weights connected from factor layers to output layer (second pass), while $\mathbf{\alpha}, \mathbf{\beta}$ represent weights from the input layer of the $k^{th}$ message passing to factor layers (first pass).\\

%[[changed by zhiwei]]
\vspace*{-6mm}

\subsubsection{Components in Factor Layers}
Now we explain in detail the different components in our model.

\textbf{Unary component}: In our message passing model, the unary component corresponds to group activity scores for an image $I$, action and pose scores for each person $I_{m}$ in frame $I$, represented as $s_{g}^{(k-1)}(I)$, $a_{h}^{(k-1)}(I_{m})$ and $r_{z}^{(k-1)}(I_{m})$ respectively. These scores are acquired from the previous step of message passing and are directly added to the output of the next message passing step. 
%$s_{g}^{(k-1)}(I)$ measures the compatibility of image $I$ with each activity label according to the global features learned by the network. $a_{h}^{(k-1)}(I_{m})$ and $r_{z}^{(k-1)}(I_{m})$ measure the compatibility of one person performing each action and each pose direction respectively given a detected person image patch. Note that we assume person detections are already given by first running a person detector in each frame.

\textbf{Group activity-action-pose factor layer $\phi$}: A group's activity is strongly correlated to the participating individuals' actions. This component for the model is used to measure the compatibility between individuals and groups. An individual's activity can be described by both pose and action, and we use this ternary scene-pose-action factor layer to capture dependencies between a person's fine-grained action (e.g.\ talking facing front-left) and the scene label for a group of people. Note that in this factor layer we used the weight sharing scheme mentioned in Sec.~\ref{sec:mp} to mimic the belief propagation. 

\textbf{Poses-all factor layer $\psi$}: Pose information is very important in understanding a group activity. For example, when all people are looking in the same direction, there is a high probability that it's a queueing scene. This component captures this global pose information for a scene. Instead of naively enumerate all combination of poses for all people, we exploit the sparsity of truly useful and frequent patterns, and simply use $T$ factor nodes for one scene label. In our experiments, we simply set $T$ to be 10.
%One naive way to capture the pose patterns for all the people in one scene is to enumerate all possible combinations of poses for $M$ people in one scene, which will lead to an exponential number of factors. However, normally the patterns for one group activity are very sparse, so we only use $T$ factor nodes to capture the most frequent patterns for each group activity label and let the network automatically learn the patterns. In our experiments, we simply set $T$ to be 10.

\subsection{Multi Step Message Passing CNN Training}
The steps of message passing depends on the structure of graphical model. In general, graphical models with loops or large number of levels will lead to more steps belief propagation for sharing local information globally. In our model, we adopt two message passing steps, as shown in Fig.~\ref{fig:label_refinement_procedure}.

\textbf{Multi-loss training:} Since the goal of our model is to recognize group activities through global features and individual actions in that group, we adopt an alternative strategy for training the model. For the $k^{th}$ message passing step, we first remove the loss layers for actions and poses to learn parameters for group activity classification alone. In this phase, there is no back-propagation on action and pose classification. Since group activity heavily depends on an individual's activity, we then fix the softmax loss layer for scene classification and learn the model for actions and poses. The trained model is used for the next message passing step. Note that in each message passing step, we exploit the benefit of the neural network structure and jointly trained the whole network.

\textbf{Learning semantic features for group activity:} Traditional convolutional neural networks mainly focus on learning features for basic classification or localization tasks. However, in our proposed message passing CNN deep model, we not only learn features, but also learn semantic high-level features for better representing group activities and interactions within the group. We explore different layers' features for this deep model, and results show that these semantic features can be used for better scene understanding and classification.

\textbf{Implementation details:} Firstly, in practice, it is not guaranteed that every frame has the same number of detections. However, the structure of neural network should be fixed. To solve this problem, denoting $M_{max}$ as the maximum number of people contained in one frame, we do a dummy-image padding when the number of people is less than $M_{max}$. Then we filter out these dummy data by de-activating neurons connected with them in related layers. Secondly, After the first message passing step, instead of directly feeding the raw scores into the next message passing step, we first normalize the pose and action scores for each person and scene scores for one frame by a softmax layer, converting to probabilities similar to belief propagation.

\section{Experiments}

Our models are implemented using the Caffe library~\cite{jia2014caffe} by defining two types of sparsely connected and weight shared inner product layers. One is from variable nodes to factor nodes, another is the reverse direction. We used TanH neurons as the non-linearity of these two layers. To examine the performance of our model, we test our model for scene classification on two datasets: (1) Collective Activity~\cite{ChoiSS2009}, (2) a nursing home dataset consisting of surveillance videos collected from a nursing home.

% Define name of model convnet -> M1; convnet + MP ->  GMconvnet;
We trained an RBF kernel SVM on features extracted from the graphical model layer after each step of message passing model.  These SVMs  are used to predict scene labels for each frame, the standard task in these datasets.

\subsection{Collective Activity Dataset}

The Collective Activity Dataset contains 44 video clips acquired using low resolution hand-held cameras. Every person is assigned one of the following five action labels:  crossing, waiting, queuing, walking and talking and one of the eight pose labels: right, front-right, front, front-left, left, back-left, back, back-right. Each frame is assigned one of the following five activities: crossing, waiting, queueing, walking, and talking. The activity category is attained by taking the majority of actions happening in one frame while ignoring the poses. We adopt the standard training test split used in~\cite{Tian_NIPS2010}.  

In the Collective Activity dataset experiment, we further concatenate the global features for a scene with AC descriptors by HOG features~\cite{Tian_NIPS2010}. We simply averaged AC descriptors features for all people and use this feature to serve as additional global information, namely this feature does not truly participated in the message passing process.  This additional global information assists in classification with the limited amount of training data available for this dataset\footnote{Scene classification accuracy solely using AlexNet is 48\%.}.

%[[changed by zhiwei]]

\begin{figure*}
	\centerline{\includegraphics[width=0.7\textwidth]{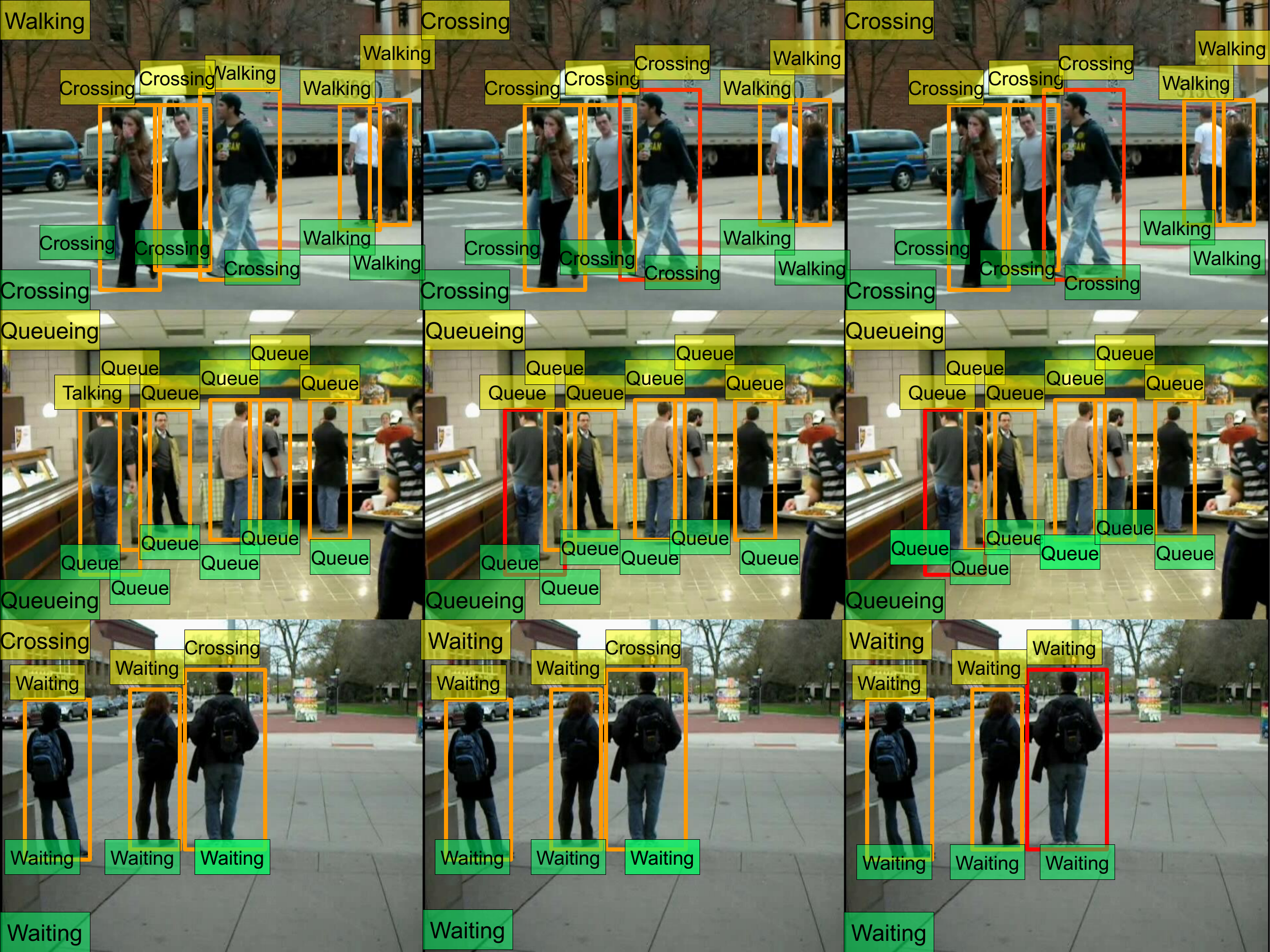}}
	\caption{Results visualization for our model. Green tags are ground truth, yellow tags are predicted labels. From left to right is without message passing, first step message passing and second step message passing}
\label{fig:vis}
\end{figure*}

We summarize the comparisons of activity classification accuracies of different methods in Tab.~\ref{table:cad}.
The current best result using spatial information in graphical model is 79.1\%, from Lan et al.~\cite{Tian_NIPS2010}, which adopted a latent max-margin method to learn graphical model with optimized structure. Our classification accuracies (the best is 80.6\%) are competitive compared with the state-of-the-art methods.  
However, the benefits of the message passing are clear. Through each step of the message passing, the factor layer effectively captured dependencies between different variables and passing messages using factor neurons results in a gain in classification accuracy. Some visualization results are shown in Fig~\ref{fig:vis}

%[[changed by zhiwei]]

% zero step message passing
% SVM + extracted features
%\begin{center}
\begin{table}[!hbp]
\begin{tabular}{|c|c|c|}
\hline
 & 1 Step MP  & 2 Steps MP  \\
\hline
Pure DL & 73.6\% &  78.4\% \\
\hline
SVM+DL Feature  & 75.1\% &  80.6\%  \\
\hline
%Latent Constituent~\cite{BorislavECCV2014} & \multicolumn{2}{|c|}{75.1\%}\\
%\hline
%Contextual model~\cite{Tian_NIPS2010} & \multicolumn{2}{|c|}{79.1\%}\\
%\hline
\end{tabular}
\begin{tabular}{|c|c|}
\hline
Latent Constituent~\cite{BorislavECCV2014} & 75.1\% \\
\hline
Contextual model~\cite{Tian_NIPS2010} & 79.1\% \\
\hline
Our Best Result & \textbf{80.6}\% \\
\hline
\end{tabular}
\caption{Scene classification accuracy on the Collective Activity Dataset.}
\label{table:cad}
\end{table}
%\end{center}
% --- only compared to spatial model

\vspace*{-5mm}

%[[ Mention current best result, note that this work is complementary and explores a deep network for doing message passing ]]

\subsection{Nursing Home Dataset}

This dataset consists 80 videos and is captured in a nursing home, including a variety of rooms such as dining rooms, corridors, etc. The 80 surveillance videos are recorded at 640 by 480 pixels at 24 frames per second, and contain a diverse set of actions and frequent cluttered scenes. This dataset contains typical actions include walking, standing, sitting, bending, squatting, and falling. For this dataset, the goal is to detect falling people, thus we assign each frame one of two activity categories: fall and non-fall. A frame is assigned ``fall" if any person falls and ``non-fall" otherwise. Note that many frames are challenging, and the falling person may be occluded by others in the scene.  We adopted a standard 2/3 and 1/3 training test split. In order to remove redundancy, we sampled 1 out of every 10 frames for training and evaluation.
Since this dataset has a large intra-class diversity within actions, we used the action primitive based detectors proposed in~\cite{Tian_eccvWS14} for more robust detection results.

Note that since this dataset has no pose attribute, we simply used one scene-action factor layer to perform the two step message passing. For the SVM classifier, only deep learning features are used. We summarize the comparisons of activity classification accuracies of different methods in Table~\ref{table:nurse}. 

%[[changed by zhiwei]]

\begin{center}
\begin{table}[!hbp]
\begin{tabular}{|c|c|c|}
\hline
 Ground Truth& Pure DL  & SVM+DL Fea.  \\
\hline
 1 Step MP & 82.5\%  & 82.3\%  \\
\hline
 2 Steps MP & 84.1\%  & 84.7\%  \\
\hline
\end{tabular}
\begin{tabular}{|c|c|c|}
\hline
 Detection & Pure DL  & SVM+DL Fea.  \\
\hline
 1 Step MP & 74.4\%  & 76.5\%  \\
\hline
 2 Steps MP & 75.6\%  & 77.3\%  \\
\hline
\end{tabular}
\caption{Classification accuracy on the nursing home dataset}.
\label{table:nurse}
\end{table}
\end{center}

%\vspace*{-\baselineskip}
\vspace*{-12mm}

The scene classification accuracy on the Nursing Home dataset by using a baseline AlexNet model is 69\%. The results on scene classification for each step also shows gains. Note that in this dataset, accuracy on the second message passing gains an increase of around 1.5\% for both pure deep learning or SVM prediction. We believe that this is due to the fact that the dataset only contains two scene labels, fall or non-fall, so scene variables are not as informative as scenes in the Collective Activity Dataset.

%[[ need to refer to the visualization figure in the text, need to improve that figure, the current text is too small on it ]]

%zhiwei]]

\section{Conclusion}
\label{sec:conclusion}
 
We have presented a deep learning model for group activity recognition which jointly captures the group activity, the individual person actions, and the interactions between them. We propose a way to combine graphical models with a deep network by mimicking the message passing process to do inference. We successfully applied this model to real scene surveillance videos and showed its' effectiveness in recognizing the activity of a group of people.

%------------------------------------------------------------------------- 

\newpage

\bibliography{egbib}
\end{document}